\PassOptionsToPackage{unicode}{hyperref}
\PassOptionsToPackage{hyphens}{url}
\PassOptionsToPackage{dvipsnames,svgnames,x11names}{xcolor}
\documentclass[
  12pt]{article}

\usepackage{amsmath,amssymb}
\usepackage{algorithm}
\usepackage{algorithmic}
\usepackage{iftex}
\usepackage{enumerate}
\usepackage{yhmath}
\newtheorem{thm}{Theorem}

\newtheorem{rem}{Remark}[section]

\ifPDFTeX
  \usepackage[T1]{fontenc}
  \usepackage[utf8]{inputenc}
  \usepackage{textcomp} 
\else 
  \usepackage{unicode-math}
  \defaultfontfeatures{Scale=MatchLowercase}
  \defaultfontfeatures[\rmfamily]{Ligatures=TeX,Scale=1}
\fi
\usepackage{lmodern}
\ifPDFTeX\else  
\fi
\IfFileExists{upquote.sty}{\usepackage{upquote}}{}
\IfFileExists{microtype.sty}{
  \usepackage[]{microtype}
  \UseMicrotypeSet[protrusion]{basicmath} 
}{}
\makeatletter
\@ifundefined{KOMAClassName}{
  \IfFileExists{parskip.sty}{%
    \usepackage{parskip}
  }{
    \setlength{\parindent}{0pt}
    \setlength{\parskip}{6pt plus 2pt minus 1pt}}
}{
  \KOMAoptions{parskip=half}}
\makeatother
\usepackage{xcolor}
\makeatletter
\ifx\paragraph\undefined\else
  \let\oldparagraph\paragraph
  \renewcommand{\paragraph}{
    \@ifstar
      \xxxParagraphStar
      \xxxParagraphNoStar
  }
  \newcommand{\xxxParagraphStar}[1]{\oldparagraph*{#1}\mbox{}}
  \newcommand{\xxxParagraphNoStar}[1]{\oldparagraph{#1}\mbox{}}
\fi
\ifx\subparagraph\undefined\else
  \let\oldsubparagraph\subparagraph
  \renewcommand{\subparagraph}{
    \@ifstar
      \xxxSubParagraphStar
      \xxxSubParagraphNoStar
  }
  \newcommand{\xxxSubParagraphStar}[1]{\oldsubparagraph*{#1}\mbox{}}
  \newcommand{\xxxSubParagraphNoStar}[1]{\oldsubparagraph{#1}\mbox{}}
\fi
\makeatother

\usepackage{longtable,booktabs,array}
\usepackage{calc} 
\usepackage{etoolbox}
\makeatletter
\patchcmd\longtable{\par}{\if@noskipsec\mbox{}\fi\par}{}{}
\makeatother
\IfFileExists{footnotehyper.sty}{\usepackage{footnotehyper}}{\usepackage{footnote}}
\makesavenoteenv{longtable}
\usepackage{graphicx}
\makeatletter
\def\maxwidth{\ifdim\Gin@nat@width>\linewidth\linewidth\else\Gin@nat@width\fi}
\def\maxheight{\ifdim\Gin@nat@height>\textheight\textheight\else\Gin@nat@height\fi}
\makeatother
\setkeys{Gin}{width=\maxwidth,height=\maxheight,keepaspectratio}
\makeatletter
\def\fps@figure{htbp}
\makeatother

\makeatletter
\@ifpackageloaded{caption}{}{\usepackage{caption}}
\AtBeginDocument{%
\ifdefined\contentsname
  \renewcommand*\contentsname{Table of contents}
\else
  \newcommand\contentsname{Table of contents}
\fi
\ifdefined\listfigurename
  \renewcommand*\listfigurename{List of Figures}
\else
  \newcommand\listfigurename{List of Figures}
\fi
\ifdefined\listtablename
  \renewcommand*\listtablename{List of Tables}
\else
  \newcommand\listtablename{List of Tables}
\fi
\ifdefined\figurename
  \renewcommand*\figurename{Figure}
\else
  \newcommand\figurename{Figure}
\fi
\ifdefined\tablename
  \renewcommand*\tablename{Table}
\else
  \newcommand\tablename{Table}
\fi
}
\@ifpackageloaded{float}{}{\usepackage{float}}
\floatstyle{ruled}
\@ifundefined{c@chapter}{\newfloat{codelisting}{h}{lop}}{\newfloat{codelisting}{h}{lop}[chapter]}
\floatname{codelisting}{Listing}

\makeatother
\makeatletter
\@ifpackageloaded{caption}{}{\usepackage{caption}}
\@ifpackageloaded{subcaption}{}{\usepackage{subcaption}}
\makeatother

\ifLuaTeX
  \usepackage{selnolig}  
\fi
\usepackage[]{natbib}
\usepackage{bookmark}

\IfFileExists{xurl.sty}{\usepackage{xurl}}{} 
\hypersetup{
  pdftitle={Title},
  pdfauthor={Author 1; Author 2},
  pdfkeywords={3 to 6 keywords, that do not appear in the title},
  colorlinks=true,
  linkcolor={blue},
  filecolor={Maroon},
  citecolor={Blue},
  urlcolor={Blue},
  pdfcreator={LaTeX via pandoc}}

\newcommand{\anon}{1}

\begin{document}

\def\spacingset#1{\renewcommand{\baselinestretch}%
{#1}\small\normalsize} \spacingset{1}


\if1\anon
{
  \title{\bf A Transfer Learning Framework  for Multilayer Networks  via Model Averaging}
  \author{Yongqin Qiu
\hspace{.2cm}\\
    International Institute of Finance, School of Management,\\ University of Science and Technology of China, Hefei 230026, China\\
    and \\
    Xinyu Zhang\thanks{Xinyu Zhang is the corresponding author (Email: xinyu@amss.ac.cn). }\\
    Academy of Mathematics and Systems Science,\\ Chinese Academy of Sciences, Beijing 100190, China\\
    International Institute of Finance, School of Management,\\ University of Science and Technology of China, Hefei 230026, China
}
  \maketitle
} \fi

\if0\anon
{
  \bigskip
  \bigskip
  \bigskip
  \begin{center}
    {\LARGE\bf Title}
\end{center}
  \medskip
} \fi

\bigskip
\begin{abstract}
Link prediction in multilayer networks is a key challenge in applications such as recommendation systems and protein-protein interaction prediction. While many techniques have been developed, most rely on assumptions about shared structures and require access to raw auxiliary data, limiting their practicality.
To address these issues, we propose a novel transfer learning framework for multilayer networks using a bi-level model averaging method.   A $K$-fold cross-validation criterion based on edges is used to automatically weight inter-layer and intra-layer candidate models. This enables the transfer of information from auxiliary layers while mitigating model uncertainty, even without prior knowledge of shared structures.
Theoretically, we prove the optimality and weight convergence of our method under mild conditions. Computationally, our framework is efficient and privacy-preserving, as it avoids raw data sharing and supports parallel processing across multiple servers.
Simulations show our method outperforms others in predictive accuracy and robustness. We further demonstrate its practical value through two real-world recommendation system applications.
\end{abstract}

\noindent%
\vfill

\newpage
\spacingset{1.8} 

\section{Introduction}\label{sec-intro}
\subsection{Background and Motivation}
In recent decades, the availability of network data has grown exponentially, sparking widespread interest in network analysis across diverse fields such as social networks \citep{lewis2012social,liu2024controlling,jin2024mixed}, biological networks \citep{nepusz2012detecting,peschel2021netcomi,xu2023covariate,zhang2024consistent}, and co-authorship networks \citep{ji2016coauthorship,ji2022co}. A fundamental challenge in network science is link prediction, which aims to predict missing or future edges based on observed ones.  For instance, suggesting new friends or connections is a core  feature of social media platforms. In biological networks, confirming protein-protein interactions through comprehensive experiments is labor-intensive and costly. Thus, link prediction can assist in pinpointing potential candidates for further experimental investigation.

Numerous methods have been developed for link prediction, including similarity-based methods, probabilistic and maximum likelihood models, dimensionality reduction techniques, and deep learning models \citep{daud2020applications,kumar2020link}. However, these approaches are primarily designed for single-layer networks.  In reality, multiple types of connections can exist between nodes. For example, Twitter captures various user interactions such as retweets, replies, mentions, and friendships.   Moreover, platforms can collaborate with other institutions to improve link prediction accuracy by integrating data from multiple sources, as demonstrated in our real-world analysis that combines Facebook and Twitter networks. These multi-faceted interactions naturally give rise to multilayer networks, where different layers share common nodes but exhibit unique edge types representing distinct relationships. A naive approach to modeling multilayer networks is to aggregate the layers into a single network, which may result in the loss of information specific to each layer due to potential heterogeneity. Conversely, modeling each layer separately ignores the shared information across networks. Therefore, developing specialized tools for multilayer network analysis is essential.

A common strategy for multilayer network analysis is joint training, which leverages shared information across layers to enhance performance. Although such methods have been applied in various scenarios, they typically suffer from several key limitations.
\begin{itemize}
	\item First, network data exist in diverse forms, such as directed or undirected networks (depending on edge directionality), weighted or unweighted networks (based on the presence of numerical edge values), and attributed or non-attributed networks (depending on whether edges possess covariates). However, most existing methods are tailored to specific network models and handle only one or a limited subset of these types, which restricts their applicability  in more diverse settings. Moreover, these model-specific approaches often  assume that different layers share similar structural properties.   When this assumption is violated, the approach may not only fail to improve estimation but could also negatively impact the results. 
	\item Second, existing methods often involve hyperparameters that require tuning, and the choice of hyperparameters can significantly influence the model's outcomes. Typically, these methods necessitate selecting the best model from a set of candidates, a process known as model selection. However, model selection may overlook valuable information contained in other models and fail to account for the uncertainties associated with the choice of models. In fact, model selection carries an underlying assumption that there exists a ``correct'' model or one that is ``more correct'' than the others. Yet, in practical applications, it is possible that all models are incorrect, but several competitive models may be equally or similarly suited for the data at hand. Even if one model outperforms others in certain scenarios, due to limited sample sizes, there is no guarantee that it will be selected.
	\item Third, these models often require access to the entire network data. In practice, however, companies and institutions may be unable to share raw network data due to commercial confidentiality and legal regulations, making these methods impractical in such contexts.
\end{itemize}

These limitations motivate the development of a unified transfer learning framework for multilayer networks. Inspired by the spirit of model averaging \citep{hu2023optimal, zhang2024prediction}, the framework offers enhanced flexibility across different network types without requiring prior knowledge of shared structures between layers. Furthermore, it mitigates model uncertainty and enables accurate estimation without access to the raw data of auxiliary networks.

	\subsection{Literature Review and Challenges}
Link prediction in multilayer networks has been extensively studied and can be broadly categorized into several approaches. One widely used category is similarity-based methods, which are grounded in the principle that two nodes are more likely to be linked if they share common neighbors \citep{yao2017link,aleta2020link,luo2022link}. Another popular approach is embedding-based methods, which aim to learn low-dimensional representations of nodes such that structurally or semantically similar nodes in the original network are mapped close to each other in the embedding space \citep{liu2017principled,du2020cross,ren2024link}. In addition, some studies model multilayer networks as high-order tensors and apply tensor decomposition techniques for link prediction \citep{lyu2023latent,aguiar2024tensor}. We restrict our review to the latent space model (LSM)—the working model used in our numerical experiments—due to its popularity, theoretical foundations, and empirical success  (It should be noted that our framework is fundamentally general and can be adapted to incorporate various model specifications based on user needs).

 LSM was originally introduced by \citet{hoff2002latent} and can be viewed as a type of embedding-based method, as it embeds each node into a low-dimensional latent space. This latent space flexibly captures several common properties of real-world networks, such as transitivity, homophily, and community structures. Regarding the application of latent space models to multilayer networks, \citet{gollini2016joint} introduced the latent space joint model, which merges the information from multiple network views. \citet{d2019latent} adopted a similar framework to study the recovery of similarities among countries by modeling the exchange of votes during several editions of the Eurovision Song Contest. \citet{salter2017latent} represented the dependence structure between network views using a multivariate Bernoulli likelihood. \citet{sosa2022latent} proposed a novel model with a key feature: A hierarchical prior distribution that allows the representation of the entire system jointly.

In the aforementioned latent space models, both the latent vectors and node effects are treated as random variables, with Bayesian approaches employed for estimation. However, the computational cost can become prohibitively high for large-scale networks. For larger networks, we focus on frequentist approaches. \citet{ma2020universal} were the first to treat latent vectors as fixed parameters and proposed using projected gradient descent for estimation. \citet{zhang2020flexible} extended this approach to multilayer networks. Their model assumes that degree heterogeneity varies across layers but that the latent vectors remain the same. \citet{macdonald2022latent} further considered heterogeneity among the latent vectors of different layers, meaning that some, but not all, structure is shared across network layers. \citet{lyu2023latent} also addressed heterogeneity by modeling each layer as independently sampled from a mixture of $m$ LSMs.

As mentioned earlier, these methods impose certain assumptions about shared structures across layers and involve a model selection process. For instance, \citet{zhang2020flexible}'s method requires specifying the dimensionality of the latent vectors, while \citet{macdonald2022latent}'s approach requires selecting a tuning parameter for the nuclear norm. Additionally, these methods necessitate access to the raw data of each layer, which can be a significant limitation in practice.

Model averaging has emerged as a promising solution to these challenges. By constructing a weighted average of predictions from all candidate models, model averaging leverages all available information and mitigates the risk of over-reliance on a single model. There are two main model averaging methods: Bayesian model averaging and frequentist model averaging. In this work, we focus on frequentist model averaging because the weights are determined solely by the data, with no assumptions about priors. Over the past two decades, frequentist model averaging has been actively developed and applied to various contexts, such as structural break models \citep{hansen2012jackknife}, mixed effects models \citep{zhang2014model}, quantile regression models \citep{lu2015jackknife, wang2023jackknife}, generalized linear models \citep{zhang2016optimal, ando2017weight}, time series models \citep{liao2021model,zhang2023optimal}, and rank-based models \citep{he2023rank, feng2024ranking}. In the context of network data, \citet{ghasemian2020stacking} empirically demonstrated the effectiveness of model averaging. \citet{li2024network}  showed that when the true model is included in the candidate set, the averaged estimator can achieve oracle-like performance; even under model misspecification,  the averaged estimator outperforms individual models. However, their method primarily focuses on undirected and unweighted networks and lacks theoretical guarantees for prediction risk optimality.

Notably, all of these methods construct candidate models based on a single dataset (or single-layer network).  In contrast to traditional model averaging methods, some recent studies have applied model averaging to transfer learning. \citet{zhang2024prediction} first proposed averaging models across different datasets to improve predictive performance for the target dataset. This approach can yield optimal predictions in large samples, even if the main model is misspecified. When the main model is correct, it automatically eliminates the influence of misspecified helper models. \citet{hu2023optimal} combined parameter information from multiple semiparametric models using samples from heterogeneous populations through a frequentist model averaging strategy. However, both methods primarily focus on reducing the uncertainty of the model for the target dataset by transferring information from different datasets. They have only one candidate model for each dataset, thereby neglecting the uncertainty of the model itself. Moreover, there is currently no frequentist model averaging method specifically designed for multilayer network data. Compared to traditional data forms, network data are more complex, as the edges do not satisfy the assumption of independence and identical distribution. In addition, the diversity of network types and cross-layer heterogeneity introduce substantial challenges for both modeling and theoretical development.

\subsection{Our Contributions and Overview}
We propose a frequentist model averaging method for transfer learning in multilayer networks. The main contributions of our work are as follows: (1) We propose a bi-level model averaging method that combines intra-layer and inter-layer models by minimizing a $K$-fold cross-validation criterion based on target layer edges. Averaging intra-layer models reduces model uncertainty, while averaging inter-layer models facilitates the transfer of information from different layers to the target layer.
(2) Our proposed method avoids assumptions about shared structures and assigns model weights in a fully data-driven manner. It is applicable to various types of networks, allowing users to choose appropriate working models based on their specific needs. (3)
We prove both weight convergence and prediction optimality of the proposed method from a non-asymptotic perspective. Specifically, if all models are non-informative but  some weighted combinations can form informative models, the estimated weights will converge to one such combination. Otherwise, when no such combination exists, the ratio of the prediction risk for all edges to the optimal risk converges to 1.  If informative models do exist, the sum of their weights will converge to 1. It is worth noting that, compared to existing literature, we relax a key condition on the rate of minimum risk for weighted combinations involving non-informative models, thereby extending the applicability of the theorems.  
(4)  Our method eliminates the need for raw data access, requiring only prediction results from auxiliary layers, thus ensuring privacy compliance.

The rest of this paper is organized as follows.  Section \ref{sec-meth} describes our proposed method in detail. In Section \ref{theory}, we develop the theoretical properties of the proposed estimator. In Section \ref{simulation}, we use simulation studies to evaluate the performance of our method. In Section \ref{real data}, we apply our method to social networks. Finally, we conclude this paper with a discussion in Section \ref{conclusion}.

\section{Methods}\label{sec-meth}

Suppose that we observe $R$ directed or undirected networks on a common set of $n$ nodes with no self-loops. The $r$th layer network is represented by an $n \times n$ symmetric adjacency matrix  $\boldsymbol{A}_r=(A_{r,ij})_{i=1,j=1}^{n,n}$.
In this paper, we propose the following general modeling framework:
\begin{equation}
	\begin{aligned}
		A_{r,ij}\overset{\text{ind}}{\sim}&f\{\cdot,\Theta_{r,ij}\} (i,j\in [n];i\ne j;r\in[R]),\\
	\end{aligned}
	\label{single layer}
\end{equation}
where $[n]$ denotes the index set $\{1,\dots, n\}$ and $f(x;\Theta)\propto \exp\{x\Theta-b(\Theta)\}$ is an exponential family distribution with natural parameter  $\Theta$. For example, if $f(x;\Theta)$ is the Gaussian  distribution $\text{Normal}(\Theta,\sigma^2)$, then   $b(\Theta)=\Theta^2/2$. By the property of exponential families, we can predict ${\mathbb E}(A_{r,ij})$ as $b'(\widehat{\Theta}_{r,ij})$, where $\widehat{\Theta}_{r,ij}$ is the estimator of ${\Theta}_{r,ij}$.
The parameter $\Theta_{r,ij}$ accommodates flexible modeling approaches. For undirected networks, one may adopt the inner product model introduced by  \citet{ma2020universal}:
\begin{equation}
	\begin{aligned}
		A_{r,ij}=&A_{r,ji}\overset{\text{ind}}{\sim}\text{Bernoulli}\left(\frac{\exp(\Theta_{r,ij})}{1+\exp(\Theta_{r,ij})}\right),\\
		\Theta_{r,ij}=&\alpha_{r,i}+\alpha_{r,j}+\ell(\boldsymbol U_{r,i},\boldsymbol U_{r,j})+\boldsymbol x^\top_{r,ij}\boldsymbol \beta_r,
	\end{aligned}
	\label{inner product}
\end{equation}
where $\boldsymbol x_{r,ij}$ are edge covariates and $\ell(\cdot,\cdot)$ is a smooth symmetric function on $\mathbb R^{d_r}\times \mathbb R^{d_r}$. The vector  $(\alpha_{r,1},\dots,\alpha_{r,n})^\top \in \mathbb R^n$ are node degree heterogeneity parameters of the $r$th layer network.  Specifically, for the $r$th layer, when all other parameters are fixed,   node $i$ is more likely to connect with other nodes as $\alpha_{r,i}$ increases.  The vector $\boldsymbol{U}_{r,i}\in \mathbb{R}^{d_r}$  is a low-dimensional latent vector associated with node $i$ and $\boldsymbol \beta_r$ is the coefficient vector for the edge covariates in the $r$th layer network. 
Beyond the inner product model, alternative approaches—such as similarity-based models and deep learning-based methods—can also estimate $\boldsymbol\Theta_{r}$.

Note that due to the presence of missing edges, we can only observe a portion of the adjacency matrix. Define $\mathcal G=\{(i,j):i,j\in [n]\text{ and } i\ne j\}$ as the set of all possible edges, and let $\mathcal G_r =\{(i,j): A_{r,ij} \text{ is observed and } i\ne j\}$ denote the set of observed edges in the $r$th layer. Additionally, let $\boldsymbol{A}_{r, \mathcal{G}'}$ denote the edges in the set $\mathcal{G}' \subset \mathcal{G}$. Assume that the first layer is the target network of interest,   and the remaining $R - 1$ layers are auxiliary networks. Our goal is to improve the prediction of missing edges in the target network with the assistance of the auxiliary networks, without imposing any assumptions on the correlation between network structures across layers. 

Although various methods can estimate  $\boldsymbol\Theta_r$, they typically require specifying tuning parameters $\nu_r$. For example, if we adopt model \eqref{inner product} and employ a convex approach for estimation, then $\nu_r$ is a penalty parameter. Alternatively, if we use a non-convex approach, then $\nu_r$ represents the latent dimension.  However, regardless of the method employed, the correct or ``more correct'' $\nu_r$ is typically unknown, even if such a value exists. Therefore, we consider multiple candidate values for $\nu_r$ with each value corresponding to a distinct model in the $r$th layer.  Let $\mathcal M_r$ be the candidate set of $\nu_r$ and $\mathcal  M=\cup_{r=1}^R \mathcal M_r$.
Without loss of generality, we set  $\mathcal M = \mathcal{M}_1 = \dots = \mathcal M_R=\{\nu_{(1)},\dots, \nu_{(M)}\}$. 
Let $\widehat{\Theta}_{rm,ij}$ be the estimator of $\Theta_{r,ij}$ with $\nu_{(m)} \in \mathcal M$,
then the prediction of $A_{1,ij}$ ($(i,j)\in\mathcal G\setminus\mathcal G_1$) is
\begin{equation}
	\widehat{A}^{(rm)}_{1,ij}=b'( {\widehat{\Theta}}_{rm,ij}).
\end{equation}
Following the idea of model averaging, we construct a prediction by averaging all possible predicted values of $\widehat{A}_{1,ij}^{(rm)}$.
Let $\mathcal W = \{\boldsymbol w=(w_1,\dots,w_{RM})^\top \in [0, 1]^{RM} :  \sum_{r=1}^{R}\sum_{m=1}^Mw_{rm}=1\}$ be the set of all possible weight vectors, where $M=|\mathcal M|$ is the number of candidate dimension of latent space, then the  weighted prediction is defined as
\begin{equation}
	\widehat{A}_{1,ij}(\boldsymbol{w})=\sum_{r=1}^{R}\sum_{m=1}^Mw_{rm}\widehat{A}_{1,ij}^{(rm)}.
\end{equation}

To determine an appropriate choice of weights, we adopt a $K$-fold cross-validation criterion, where $1<K\le|\mathcal G_1|$.  Specifically, we randomly divide $\mathcal G_{1}$ into $K$ mutually exclusive groups $\mathcal G_{1}^{(1)},\dots,\mathcal G_{1}^{(K)}$. Then the $K$-fold cross-validation based weight choice criterion is defined as 
\begin{equation}
	CV(\boldsymbol w)=\sum^K_{k=1}\sum_{(i,j)\in \mathcal G_1^{(k)}}\left(A_{1,ij}-\widehat{A}^{(-k)}_{1,ij}\left(\boldsymbol w\right)\right)^2,
\end{equation}
where 
\begin{equation}
	\widehat{A}^{(-k)}_{1,ij}\left(\boldsymbol w\right) =
	\sum_{m=1}^Mw_{1m}\widehat{A}_{1,ij}^{(1m,-k)}+ \sum_{r=2}^{R}\sum_{m=1}^Mw_{rm}\widehat{A}_{1,ij}^{(rm)},
	\label{weighted combination}
\end{equation} and $\widehat{A}_{1,ij}^{(1m,-k)}$ is defined similarly to  $\widehat{A}_{1,ij}^{(1m)}$ but is estimated using only the edges in the set difference $\mathcal G_1\setminus \mathcal G_1^{(k)}$. 
The weight vector can be obtained by solving the following constrained optimization problem:
\begin{equation}
	\widehat{\boldsymbol w} = \arg\min_{\boldsymbol w \in \mathcal W}CV(\boldsymbol w).
	\label{cv criteria}
\end{equation}
We summarize our procedure in Algorithm 1.

\begin{algorithm}[H]
	\caption{Transfer-MA} 
	\hspace*{0.02in} {\bf Input:} 
	$\{\boldsymbol{A}_{r,\mathcal G_r}\}^R_{r=1}$;  $\mathcal M$
	\begin{algorithmic}[1]
		\FOR {$r=1,2,\dots,R$}
		\FOR {$m = 1,\dots,M$}
		\STATE Estimate $\widehat{\boldsymbol \Theta}_{rm}=(\widehat{ \Theta}_{rm,ij})^{n,n}_{i=1,j=1}$ using the edges $\mathcal G_r$
		\ENDFOR
		\ENDFOR\\
		\FOR {$k=1,2,\dots,K$}
		\FOR {$m = 1,\dots,M$}
		\STATE Estimate $\widehat{\boldsymbol \Theta}_{1m}^{(-k)}=(\widehat{ \Theta}^{(-k)}_{1m,ij})^{n,n}_{i=1,j=1}$  using the edges $\mathcal G_1\setminus\mathcal G_1^{(k)}$
		\ENDFOR\\
		\ENDFOR\\
		\STATE Construct the weighted combination by \eqref{weighted combination}
		\STATE Select the weight vector $\widehat{\boldsymbol w}$ by minimizing \eqref{cv criteria}
		\STATE Return $	\widehat{\boldsymbol A}_{1}(\widehat{\boldsymbol{w}})=\sum_{m=1}^M\sum_{r=1}^{R}\widehat{w}_{rm}\widehat{\boldsymbol A}_{1}^{(rm)}$
	\end{algorithmic}
	\label{alg2}
\end{algorithm}
It is worth noting that the estimation for each layer can be performed in parallel across multiple servers. On the one hand, this reduces computational time; on the other hand, it means that the raw data from the auxiliary layers are not required; only the predicted probabilities need to be provided. Therefore, estimation can still be conducted even under privacy constraints.

\section{Theoretical Results}\label{theory}
In this section, we study the theoretical properties of the proposed estimator.  
Let $a_n\asymp b_n$ and $a_n \lesssim b_n$ denote  $a_n/b_n\rightarrow c$ and $|a_n/b_n|\le c<\infty$ for some constant $c > 0$ respectively.  For any vector $\boldsymbol v=(v_1,\dots,v_p) \in \mathbb R^p$, let $\|\boldsymbol{v}\|_1$ and $\|\boldsymbol{v}\|_2$ denote its $\ell_1$ and $\ell_2$ norms.  For any matrix $\boldsymbol V = (V_{i,j})^{p_1,p_2}
_{i=1,j=1} \in \mathbb R^{p_1\times p_2}$, let $\|\boldsymbol V\|_{op}$  and $\|\boldsymbol{V}\|_F$ denote its spectral norm and Frobenius norm.  For any arbitrary symmetric matrix $\boldsymbol M \in \mathbb R^{p\times p}$,
define  $\lambda_{\max}(\boldsymbol M)$ and $\lambda_{\min}(\boldsymbol M )$ to be the maximum and minimum eigenvalues of $\boldsymbol M$. 
Given an index set $\mathcal A$, let $\boldsymbol v_{\mathcal A}$ denote the subvector
of $\boldsymbol v$ formed by components in $\mathcal A$. Similarly, let $\boldsymbol V_{\mathcal A}$
be the submatrix of $\boldsymbol V$ formed by columns with indices in $\mathcal A$.

\subsection{Theoretical Properties under the Existence of Informative Models}
We first consider the scenario in which some of the candidate models are capable of accurately capturing the true data-generating process (DGP) underlying the adjacency matrix $\boldsymbol{A}_1$. 
To establish the statistical properties of the proposed estimator, we assume the following conditions.

\begin{enumerate}[(C1)]
	\item Let $\epsilon_{1,ij} = A_{1,ij}-{\mathbb E}(A_{1,ij})$ denote the error term in the first layer. We assume that $\epsilon_{1,ij}$ are independent random variables with uniformly bounded variances, i.e., there exists a constant $C>0$ such that   $\text{Var}(\epsilon_{1,ij})\le C$ for all  $i,j\in[n]$.
	\item $b'(\cdot)$ is Lipschitz continuous. That is 
	$$
	|b'(x)-b'(y)|\le C_L|x-y| \;\text{for any } x,y\in \mathbb R,
	$$
	where $C_L>0$ is a constant.
	\item For each $r \in [R]$ and $m \in [M]$, there exists a value $\tilde{\boldsymbol\Theta}^*_{rm} \in \mathcal{F}_{rm}$ such that:  
		(i) $\|\widehat{\boldsymbol\Theta}_{rm} - \tilde{\boldsymbol\Theta}^*_{rm}\|_F^2/n^2 \lesssim \psi(\pi_r, n)$ with probability at least $1 - \phi(\pi_r, n)$, and  
		(ii) $|\tilde{\boldsymbol\Theta}^*_{rm,ij} - {{\mathbb E}}({A}_{1,ij})| \le C_1$ for some constant $C_1$,  
		where $\pi_r$ denotes the observation rate of edges in the $r$th layer, and $\mathcal{F}_{rm}$ denotes the feasible parameter space, which depends on the working model and the prespecified value $\nu_{(m)}$. 
		The functions $\psi(\pi, n)$ and $\phi(\pi, n)$ are positive and decreasing sequences that converge to $0$ as $n \to +\infty$, and satisfy the following conditions:  
		(a) $\lim_{n \to +\infty} n^2 \psi(1, n)R^{-1}M^{-1} = +\infty$;
		(b) $\psi(\pi, n) \asymp \psi(1, n)$ and $\phi(\pi,n)\asymp\phi(1,n)$ for all $0 < c\le\pi \le 1$, where $c$ is a constant; (c) $\lim_{n\rightarrow +\infty}(RM+K)\phi(1,n)= 0$.
	
\end{enumerate}
\begin{rem}
	Conditions (C1) and (C2) are satisfied by many popular distribution families, such as Gaussian and binomial. 
	Condition (C3)(i) assumes the convergence of the estimators. Regardless of the underlying DGP of $\boldsymbol{A}_r$, the estimator converges to a pseudo-true parameter.  This is a standard assumption in the literature, including works such as \citet{ando2014model}, \citet{ando2017weight}, and \citet{zhang2024prediction}. In the context of network data, \citet{ma2020universal} established similar results for inner product models under the convex approach with $\psi(1,n) = 1/n$ and $\phi(1,n)=n^{-c}$ ($c$ is a  positive constant). Condition (C3)(ii) concerns the boundedness of the approximation error.
	It is worth noting that condition (C3) does not require a specific form for the working model; it only requires that the estimator is convergent. Therefore, it is applicable to a wide variety of methods, including deep learning-based approaches (Although few studies formally establish convergence for deep learning methods in network data, we expect it to hold in practice). 
	
\end{rem}
	Let $h_{rm}=\sqrt{\sum_{(i,j)\in \mathcal G_1}[{\mathbb E}(A_{1,ij})-b'(\tilde{\Theta}^*_{rm,ij})]^2}$ denote the distance between the pseudo-true parameters and ${{\mathbb E}}(\boldsymbol{A}_1)$ evaluated over the observed edges. A smaller value of $h_{rm}$ indicates that the $(r,m)$th model more accurately specifies the target layer. If $h^2_{rm}/n^2$ is of the same order or smaller than the convergence rate  $\psi(1,n)$ of the estimator, then the $(r,m)$th model is regarded as informative. Define the informative set as 
	$${\mathcal S}=\{(r,m): r\in [R], m\in[M], h_{rm} \lesssim n\psi^{1/2}(1,n)\}.$$ To ensure that $\mathcal S$ is not empty, the working model and tuning parameters must be chosen carefully to adequately capture the true DGP.

	Let $\mathcal W_{\mathcal S^c}=\{\boldsymbol w\in \mathcal W:\boldsymbol 1^{\top}_{RM-|\mathcal S|}\boldsymbol w_{\mathcal S^c}=1\} $ denote the set of weight vectors that assign all weight exclusively to the non-informative models.
	For all $\boldsymbol w \in \mathcal W_{\mathcal S^c}$, define the approximation error of the weighted non-informative models as
	$${h}_{\mathcal S^c}(\boldsymbol w) = \sqrt{ \sum_{(i,j)\in \mathcal G_1}[{\mathbb E}(A_{1,ij})-\sum_{(r,m)\in \mathcal S^c}{w}_{rm}b'(\tilde{\Theta}_{rm,ij}^{*})]^2},$$
	which measures the distance between ${\mathbb E}(\boldsymbol A_{1})$ and the weighted pseudo-true parameters of the non-informative models. Define the minimum such approximation error as
	$
	\underline{h}_n = \inf\limits_{\boldsymbol w\in \mathcal W_{\mathcal S^c} }{h}_{\mathcal S^c}(\boldsymbol w)
	$
	and let $\underline{\mathcal W}_{\mathcal S^c}=\{\boldsymbol w\in{\mathcal W}_{\mathcal S^c}: {h}_{\mathcal S^c}(\boldsymbol w)\lesssim n\psi^{1/2}(1,n) \}$ denote the set of weight vectors that combine non-informative models to form informative models.

\begin{thm}Under conditions (C1)-(C3),
	if $\lim_{n\rightarrow +\infty}\underline{h}_nn^{-1}\psi^{-1/2}(1,n)= +\infty$, then 
		$${\mathbb P}\left(\|\widehat{\boldsymbol w}_{\mathcal S^c}\|_2\lesssim \frac{n\psi^{1/2}(1,n)}{\underline{h}_n}\right)\ge 1-(RM+K)\phi(C_K,n)-{c_0RM}{n^{-2}\psi^{-1}(C_K,n)},$$
		where $C_K=(K-1)\pi_{\min}/K$, $\pi_{\min}=\min_{r\in[R]}\pi_r$ and $c_0$ is a positive constant. Consequently, $\widehat{\tau}=\boldsymbol 1_{|\mathcal S|}^\top\widehat{\boldsymbol w}_{\mathcal S}\rightarrow 1$ in probability  as $n\rightarrow +\infty$.\\
\end{thm}
\begin{rem}
	The condition $\lim_{n\rightarrow +\infty}\underline{h}_nn^{-1}\psi^{-1/2}(1,n)= +\infty$ implies that the non-informative models cannot approximate the true DGP of the target layer through any weighted combination. When this condition and conditions (C1)–(C3) hold, Theorem 1 shows that the weights of non-informative models converge to 0 at the rate of ${n\psi^{1/2}(1,n)}{\underline{h}^{-1}_n}$. Equivalently, the sum of the weights assigned to the informative models converges to one at the same rate. This result implies that the model averaging procedure automatically assigns weight only to models that adequately capture the underlying DGP of the target layer, while effectively excluding poorly performing models and thus avoiding negative transfer. In the special case where $\mathcal S$ contains only a single model—that is, when exactly one model is informative—our method is capable of accurately identifying this unique informative model.  Therefore, Theorem 1 can be interpreted as a type of consistency property in model selection. 
\end{rem}
	Theorem 1 establishes weight convergence in scenarios where non-informative models cannot be combined to form an informative model. In contrast, when the condition $\lim_{n\rightarrow +\infty}\underline{h}_nn^{-1}\psi^{-1/2}(1,n)= +\infty$ is violated—i.e.,   $\lim_{n\rightarrow +\infty}\underline{h}_nn^{-1}\psi^{-1/2}(1,n)< +\infty$ or equivalently $\underline{h}_n\lesssim n\psi^{1/2}(1,n)$, there exist weight combinations such that the resulting weighted models exhibit the same property as the informative models in $\mathcal{S}$. These new models, formed by weighting non-informative ones, can therefore be regarded as informative. The next theorem demonstrates that, in this case, the estimated weights will converge to one of the combinations that yield such informative weighted models.
\begin{thm}
	Under conditions (C1)-(C3), if $\lim_{n\rightarrow +\infty}\underline{h}_nn^{-1}\psi^{-1/2}(1,n)< +\infty$, then $\underline{\mathcal W}_{\mathcal S^c}$ is not empty, and
	there exists a positive sequence $\eta_n\rightarrow 0$, such that
	$${\mathbb P}\left(\inf_{\boldsymbol w\in \underline{\mathcal W}_{\mathcal S^c}}\|\widehat{\boldsymbol w}_{\mathcal S^c}/\|\widehat{\boldsymbol w}_{\mathcal S^c}\|_1-\boldsymbol w_{\mathcal S^c}\|_2\lesssim \eta_n\Big|\|\widehat{\boldsymbol w}_{\mathcal S^c}\|_1\ge c_1\right)
	\ge 1-(RM+K)\phi(C_K,n)-{c_0RM}{n^{-2}\psi^{-1}(C_K,n)},$$
	where $c_1$ is a positive constants.  Consequently, provided $\|\widehat{\boldsymbol w}_{\mathcal S^c}\|_1\ge c_1$, we have $\inf_{\boldsymbol w\in \underline{\mathcal W}_{\mathcal S^c}}\|\widehat{\boldsymbol w}_{\mathcal S^c}/\|\widehat{\boldsymbol w}_{\mathcal S^c}\|_1-\boldsymbol w_{\mathcal S^c}\|_2\rightarrow 0$ in probability as $n\rightarrow +\infty$.  
\end{thm}
\begin{rem}
	The condition $\lim_{n\rightarrow +\infty}\underline{h}_nn^{-1}\psi^{-1/2}(1,n)< +\infty$ shows that
    the models in $\mathcal S^c$ can approximate the target layer through a weighted combination. Such a weighted model can be regarded as an informative model.  The set $\mathcal W_{\mathcal S^c}$ 
	comprises all such  weight combinations.  Theorem 2 states that if $\widehat{\boldsymbol w}_{\mathcal S^c}$ is bounded away from $\boldsymbol 0$,  then the proportions it assigns to models in $\mathcal S^c$ will converge to those of one combination in $\mathcal W_{\mathcal S^c}$. In particular, if all models are non-informative, then $\widehat{\boldsymbol w}_{\mathcal S^c}=\widehat{\boldsymbol w}$, and the estimated weight vector will converge to an ideal set of weights that combines all candidate models to form informative models.
\end{rem}

However, in practice, since the underlying DGP is typically unknown, making it uncommon for a working model to achieve good approximation. Therefore, in the next subsection, we consider the case where all models are non-informative. 

\subsection{Theoretical Properties under the Absence of Informative Models}
In this section, we investigate the predictive performance in the absence of informative models, including the prediction risks for both observed and missing edges. Define the risk function as $\mathcal R(\boldsymbol w)=\|\widehat{\boldsymbol A}_{1}(\boldsymbol{w})-{\mathbb E}(\boldsymbol A_1)\|_F^2$, and let
$\mathcal R^*(\boldsymbol w)=\|\tilde{\boldsymbol A}_{1}(\boldsymbol{w})-{\mathbb E}(\boldsymbol A_1)\|_F^2$ denote the risk function computed based on the pseudo-true values, where $\tilde{\boldsymbol A}_{1}(\boldsymbol{w})=\sum_{r=1}^{R}\sum_{m=1}^Mw_{rm}b'(\tilde{\boldsymbol\Theta}^*_{rm})$. Let $\xi_n = \inf_{\boldsymbol w\in \mathcal W}\mathcal R^*(\boldsymbol w)$ be the minimum risk  if the pseudo-true parameters were known, though these parameters cannot be obtained in practice.  To study the theoretical properties of predictions for all edges, we further assume the following two conditions:	

\begin{enumerate}[(C4)]
	\item
	$
		\lim_{n\rightarrow +\infty}{|RM\log(\xi_n/n^2)n^2\psi(1,n)|}/{\xi_n}= 0.
		$
	
\end{enumerate}

\begin{enumerate}[(C5)]
	\item For any $\boldsymbol w \in \mathcal W$, there exists a corresponding $\sigma_{\boldsymbol w}$ such that:
		$$\lim_{n\rightarrow +\infty}\sup_{\boldsymbol w\in \mathcal W}\frac{\frac{1}{|\mathcal G_1|}\sum_{(i,j)\in \mathcal G_1}\left\{\sum_{r=1}^{R}\sum_{m=1}^{M}w_{rm}\left[b'(\tilde{\Theta}^*_{rm,ij})-{\mathbb E}(A_{r',ij})\right]\right\}^2-\sigma_{\boldsymbol w}}{\sigma_{\boldsymbol w}}=0$$ 
		and 
		$$\lim_{n\rightarrow +\infty}\sup_{\boldsymbol w\in \mathcal W}\frac{\frac{1}{|\mathcal G\setminus\mathcal G_1|}\sum_{(i,j)\in \mathcal G\setminus\mathcal G_1}\left\{\sum_{r=1}^{R}\sum_{m=1}^{M}w_{rm}\left[b'(\tilde{\Theta}^*_{rm,ij})-{\mathbb E}(A_{r',ij})\right]\right\}^2- \sigma_{\boldsymbol w}}{\sigma_{\boldsymbol w}}=0.$$
\end{enumerate}

\begin{rem} Condition (C4) implies that $\lim_{n\rightarrow +\infty}n^2\psi(1,n)/\xi_n = 0$, indicating that the candidate models are sufficiently far from the true DGP. 
		It is worth noting that condition (C4) is weaker than similar conditions in existing studies \citep{ando2014model, ando2017weight, zhang2024prediction}.
		For ease of comparison, let $N = n(n-1)$ and assume $RM = O(1)$. Defining the average risk per edge as $\tilde{\xi}_N = \xi_n / N$, standard theoretical results in the literature typically require that $\tilde{\xi}_N$ satisfy $\tilde{\xi}_N N^{1/2} \to +\infty$,  which restricts its convergence rate to be no faster than $N^{-1/2}$.
		However, our condition (C4) relaxes this requirement. For example, let $\tilde{\xi}_N = N^{-1+\delta_1}$ with $0<\delta_1<1/2$ and $\psi(1,n) = N^{-1+\delta_2}$ with $\delta_2 < \delta_1$. Under these specifications, $\tilde{\xi}_N$ converges faster than  $N^{-1/2}$ while still satisfying condition (C4), as
			$$
			{|RM\log(\xi_n/n^2)n^2\psi(1,n)|}/{\xi_n}\asymp {[(1-\delta_1)\log(N)]}/N^{\delta_1-\delta_2}\rightarrow +\infty.
			$$
		Condition (C5) imposes a constraint on the signal strength of the pseudo-true values, ensuring that the average prediction strength for both observed and missing edges based on these values remains comparable. A detailed discussion on the plausibility of condition (C5) is provided in the supplementary material. 
	
Combining conditions (C4) and (C5), we conclude that all candidate models are non-informative and cannot be combined to form an informative model. To see this, suppose instead that the candidate models could be combined into an informative model. Then we would have
	$\lim_{n\rightarrow +\infty}n^2\psi(1,n)/\underline{h}_n > 0.$ Under condition (C5), this implies
	$\lim_{n\rightarrow +\infty}n^2\psi(1,n)/\xi_n > 0$, which contradicts condition (C4).

\end{rem}

\begin{thm}
		Under conditions (C1)-(C5), we have
		$$
		{\mathbb P}\left(\left|\frac{\mathcal R(\widehat{\boldsymbol w})}{ \xi_n}-1\right|\lesssim \frac{n\psi^{1/2}(1,n) }{\sqrt{\xi_n}}+\beta_n\right)\ge 1-(RM+K)\phi(C_K,n)-{c_0RM}{n^{-2}\psi^{-1}(C_K,n)},
		$$
		where $\beta_n$ is a positive sequence converging to 0. This implies that ${\mathcal R(\widehat{\boldsymbol w})}/{ \xi_n}\rightarrow 1$ in probability  as $n\rightarrow +\infty$.
\end{thm}
\begin{rem}
	Theorem 2 establishes that when the target layer cannot be well approximated by any weighted combination of all candidate models,  the ratio of the prediction risk for all edges to the optimal risk converges to 1. 
\end{rem}

\section{Simulation}\label{simulation}
In this section, we conduct simulation experiments to evaluate the numerical performance of our proposed method. We use the following inner product model as our working model: 
\begin{equation}
	\Theta_{r,ij}=\alpha_{r,i}+\alpha_{r,j}+\boldsymbol{U}_{r,i}^{\top}\boldsymbol{\Lambda}_r\boldsymbol{U}_{r,j},
	\label{flexible model}	
\end{equation} 
where $\boldsymbol \Lambda_r \in \mathbb R^{d_r\times d_r}  (r \in [R])$ is a layer-specific connection matrix, allowing general interactions between different dimensions of the latent vectors  $\boldsymbol{U}_r=[\boldsymbol{U}_{r,1},\dots, \boldsymbol{U}_{r,n}]^\top \in \mathbb R^{n\times d_r}$.  When $f(x;\Theta)\propto \exp(x\Theta-\log(1+\exp(\Theta)))$, it coincides with the model of \citet{zhang2020flexible}. 
Note that additional constraints on the parameters are required for the model (\ref{flexible model}) to be identifiable. Proposition 1 in the supplementary material states the identifiability conditions.  Let $\{\boldsymbol U_r^*,\boldsymbol \alpha^*_r,\boldsymbol \Lambda^*_r\}$ denote the true parameters of $\{\boldsymbol U_r,\boldsymbol \alpha_r,\boldsymbol \Lambda_r\}$. We employ the maximum likelihood estimator (MLE) to estimate the unknown parameters. The negative conditional log-likelihood for the $r$th layer is given by:
\begin{equation}
	\ell(\boldsymbol{\alpha}_r,\boldsymbol{U}_r,\boldsymbol{\Lambda}_r)\propto -\sum_{(i,j)\in \mathcal G_r}\log f(A_{r,ij},\Theta_{r,ij}).
\end{equation} 
We use the projected gradient descent algorithm (the procedure is summarized in Algorithm A.1 in the supplementary material) to obtain the maximum likelihood estimator. In this algorithm,  $\nu_{(m)}=d_{(m)}$ denotes the candidate latent dimension, and $\mathcal M = \{d_{(1)},\dots, d_{(M)}\}$ denotes the set of candidate dimensions.
\subsection{Experimental Setup}

\textbf{Example 1:} In this example, we fix $d_1=\dots=d_R=2$. 
For the target layer ($r=1$),  we generate the true parameters following the process described by \citet{zhang2020flexible}.
\begin{itemize}
	\item Generate of $\boldsymbol U^*_1$: (1) We generate $\tilde{\boldsymbol U}^*_1$ by drawing each element independently from $\text{Normal}(0,1)$.  (2) Set ${\boldsymbol U}^*_1 = \tilde{\boldsymbol U}^*_1\boldsymbol J_n$. (3) We normalize $\boldsymbol U^*_1$ such that $\boldsymbol U^{*\top}_1\boldsymbol U^*_1$. The steps (2) and (3) are used to make $\boldsymbol U^*_1$ satisfy the identifiability conditions.
	\item Generate of $\boldsymbol \alpha^*_1$: For $i\in [n]$, we set $\alpha^*_{1,i}\overset{\text{iid}}{\sim}\text{Uniform}[-2,-1]$.
	\item Generate of $\boldsymbol \Lambda^*_1$: Let $\boldsymbol \Lambda^*_1 = \text{diag}(\lambda_{1,1},\dots,\lambda_{1,d_1})$, where $\lambda^*_{1,l}\overset{\text{iid}}{\sim}\text{Uniform}[-1,-0.5]$ for $l\in [d_1]$.
\end{itemize}

For $r = 2,\dots, R$, $\boldsymbol \alpha^*_r$ and $\boldsymbol\Lambda^*_r$ are generated following the same procedure as for the target layer. The generation of $\boldsymbol U^*_r$ differs as follows: We first set $\tilde{U}^*_{r,il} = \tilde{U}^*_{1,il} + \text{Uniform}(-\sigma,\sigma)$ for $i\in[n]$ and $l \in [d_r]$, where 
$\sigma$ controls the difference between the target layer and the auxiliary layers.  Then, we generate $\boldsymbol U^*_r$ as in steps (2) and (3) for the target layer. Notably, when $\sigma=0$, the generation process for the multilayer network is identical to that of \cite{zhang2020flexible}.  For each layer, we generate  $	\Theta^*_{r,ij}=\alpha^*_{r,i}+\alpha^*_{r,j}+\boldsymbol{U}_{r,i}^{*\top}\boldsymbol{\Lambda}^*_r\boldsymbol{U}^*_{r,j}$ and consider both continuous and binary edges. The adjacency entries $A_{r,ij}$ are generated according to the following schemes:
\begin{itemize}
	\item Logistic type: In this case, $A_{r,ij}$ follow a Bernoulli distribution with the success probability given by the logistic function:  $A_{r,ij}\sim \text{Bernoulli}\left(\exp(\Theta^*_{r,ij})/(1+\exp(\Theta^*_{r,ij}))\right)$.
	\item Gaussian type: In this case,  $A_{r,ij}=\Theta^*_{r,ij}+v_{r,ij}$, where $v_{r,ij}\overset{\text{iid}}{\sim} \text{Normal}(0,20)$. 
\end{itemize}

We set $n=200,R=4,\sigma = 3$, $K=10$ and  $\mathcal M=\{2\}$ as the baseline configuration and separately change the values of $n,R,\sigma$ and $\mathcal M$. Specifically, we vary (a) $n \in \{160,180,200,220,240\}$, (b) $R \in \{4,5,6,7,8\}$, (c) $\sigma\in\{0,1,2,3,4,5\}$, (d) $K\in\{5,10,20,50,100\}$ and (e) $\mathcal M\in \left\{\{1\},\{2\},\{3\},\{1,2,3\}\right\}$.

\textbf{Example 2:} In this example, we evaluate the performance of the proposed method when heterogeneity in the auxiliary layer arises from both noise in the latent vectors and differences in latent dimensionality. The generation process for the auxiliary layers follows a similar procedure to Example 1, except for the generation of 
$\tilde{\boldsymbol U}^*_r$. Specifically,  we set  $\tilde{ U}^*_{r,il} = \tilde{ U}^*_{1,il} + \text{Uniform}(-3,3)$ if $i\le \min({d_r,d_1})$ and $\tilde{ U}^*_{r,il}=\text{Normal}(0,1)$ otherwise.  We set $n=200,R=4$,  $d_1=3,d_2=1,d_3=2,d_4=4$ and $K=10$. Here, the second and third layers have smaller latent dimensions than the target layer, while the fourth layer has a larger dimension. We consider both Gaussian type and logistic type networks, and set the candidate sets as $\mathcal{M}\in\{\{1\},\{2\},\{3\},\{4\},\{5\},\{1,2,3,4,5\}\}$.

\textbf{Example 3:} In this example, we examine weight convergence in a setting where non-informative models cannot be combined to form an informative model.  Specifically, we set $R=7$, $d_1=\dots=d_R=2$. The latent vectors $\boldsymbol U^*_1$ are generated following the same procedure as in Example 1. We fix 
$\boldsymbol \alpha^*_1=\boldsymbol 0_{n}$ and $\boldsymbol \Lambda_1^*=\text{diag}(-1,\dots,-1)$. Then, $\boldsymbol \Theta^*_r$ are generated as follows, for \( i \ne j \in [n] \):
$$
\begin{aligned}
		\Theta^*_{1,ij}&=\alpha^*_{1,i}+\alpha^*_{1,j}+\boldsymbol{U}_{1,i}^{*\top}\boldsymbol{\Lambda}^*_1\boldsymbol{U}^*_{1,j},
	 \Theta^*_{r,ij}=\Theta^*_{1,ij}+5/n^{0.6} \text{ for }  r=2,3,\\
  \Theta^*_{r,ij}&=\Theta^*_{1,ij}+5/n^{0.3} \text{ for }  r=4,5,\text{ and }
	  \Theta^*_{r,ij}=\Theta^*_{1,ij}+5 \text{ for }  r=6,7.
\end{aligned} 
$$
  We focus on Gaussian type networks, with $\mathcal M=\{2\}$ and $n\in\{200,300,400\}$. Under this setup, only the first three layers are informative, while the remaining layers are non-informative.

\textbf{Example 4:} In this example, we examine weight convergence in a setting where non-informative models can be combined to form an informative model.  Specifically, we set $R=3$, $d_1=\dots=d_R=2$. The generation of  $\boldsymbol U^*_1$, $\boldsymbol{\alpha}^*_1$ and $\boldsymbol{\Lambda}^*_1$ follows the same procedure as in Example 3. For $i \ne j \in [n]$, we set \[
	\begin{aligned}
		\Theta^*_{1,ij} &= \alpha^*_{1,i} + \alpha^*_{1,j} + \boldsymbol{U}^{*\top}_{1,i} \boldsymbol{\Lambda}^*_1 \boldsymbol{U}^*_{1,j}, 
		\Theta^*_{2,ij} = \Theta^*_{1,ij} + 5, \text{ and }
		\Theta^*_{3,ij} = \Theta^*_{1,ij} - 5.
	\end{aligned}
	\]
 We focus on Gaussian type networks, with $\mathcal M=\{2\}$ and $n\in\{200,300,400\}$.
  Under this setup, the second and third layers are individually non-informative, but their $1:1$ weighted combination yields an informative model.

\textbf{Example 5:} In this example, we evaluate the performance of the proposed method when the working model is misspecified relative to the true DGP of the target layer. We generate the true parameters as in the baseline configuration and set $A_{r,ij}\sim \Phi\left(\Theta^*_{r,ij}\right)$ where $\Phi(\cdot)$ is the cumulative distribution function of the standard normal distribution. However, we estimate the parameters using a logistic model; therefore, the working model is misspecified in this example.

For each example, we assume a missing rate of 25\% for each layer. Specifically, for each layer, we randomly select 25\% of the edges to be missing.
To better gauge the performance of our proposed method, we compare these methods in our simulation experiments. 
\begin{itemize}
	\item Transfer-MA: Our proposed transfer learning method based on model averaging.
	\item Transfer-SimpMA: Transfer learning using simple averaging, i.e., setting $w_{11}=\dots=w_{RM}=1/RM$. The candidate set $\mathcal M$ is the same as in Transfer-MA.
	\item Target-Only:  Estimation using only the target layer, with one  $d\in \mathcal M$ specified at a time.
	\item MultiNeSS: A latent space model for multiple networks with shared structure, as proposed by \citet{macdonald2022latent}. This method does not require specifying the latent vector dimensions and uses cross-validation to select the optimal model.
	\item FMLSM: A flexible latent space model for multilayer networks, as proposed by \citet{zhang2020flexible}. One  $d\in \mathcal M$ is specified at a time.
\end{itemize}

\subsection{Simulation Results}
\begin{figure}[H]
	\centering
	\includegraphics[width=6in,height=\textheight]{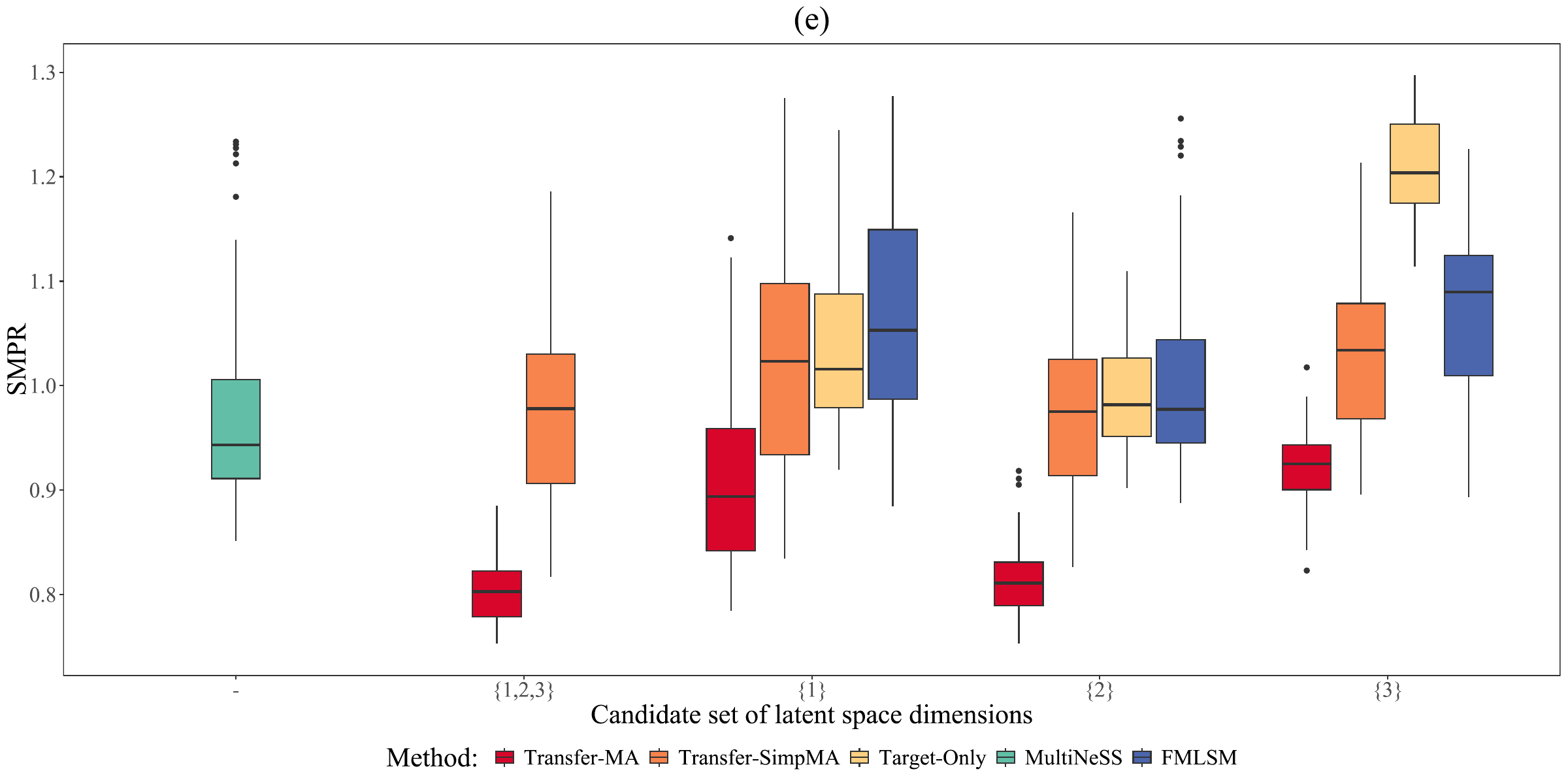}
	\caption{Results of Example 1: Case (e) (Gaussian Type)} 
	\label{gaussian example base1}
\end{figure}

To evaluate the performance of the methods described above, we calculate the square of mean prediction risk (SMPR) as follows: $\text{SMPR}=\sqrt{\sum_{(i,j)\in \mathcal G\setminus\mathcal G_1}\left(\widehat{A}_{1,ij}(\boldsymbol w)-{\mathbb E}(A_{1,ij})\right)^2}$. The evaluation procedure is repeated 100 times. For Example 1-2, our analysis primarily concentrates on Gaussian type, with analogous findings for logistic type detailed in the supplementary material.  Figures \ref{gaussian example1} to \ref{gaussian example base1} present the results for Example 1. For cases (a)-(c), line graphs are used to display the median of the 100 repeated trials, highlighting the trends.  In case (d), we compare the boxplots of Transfer-MA across different values of  $K$. In case (e), we present boxplots for different values of $\mathcal M$. Since MultiNeSS does not require the input of latent space dimensions and selects the model using cross-validation, we represent this method by a dash (``-''). Additionally, since Target-Only and FMLSM are constrained to specifying a single latent space dimension at a time, only the results for Transfer-MA and Transfer-SimpMA are presented when $\mathcal{M}=\{1,2,3\}$.

Several observations can be made from the figures:
Case (a): When the latent vectors across all layers are the same (i.e., $\sigma = 0$), FMLSM performs the best. This is not surprising, as this scenario satisfies the model's assumptions.  As $\sigma$ increases,  the performance of all models declines due to the diminishing transferable information across layers. However, our proposed method consistently outperforms the others, highlighting its effectiveness.    Notably, for large $\sigma$, other methods underperform Target-Only, indicating a negative effect on the prediction of the target layer. However, Transfer-MA consistently improves performance, demonstrating its robustness.
(2) Case (b): As the network size increases (i.e., when 
$n$ increases), prediction performance improves for all methods.
(3) Case (c): As the number of layers increases, the amount of information available for prediction also increases, leading to an improvement in the prediction performance of all methods.
(4) Case (d): The  results for different values of $K$ are similar, indicating that the prediction performance of our method is not sensitive to the choice of $K$.
(5) Case (e): For $\mathcal M$ with a single candidate dimension, our method essentially averages the intra-layer models. Performance is optimal for $\mathcal M = \{2\}$, as $d_r = 2$ matches the true dimension. For $\mathcal M = \{3\}$, performance slightly declines due to limited sample size. For $\mathcal M = \{1\}$, Transfer-MA, Transfer-SimpMA, Target-Only, and FMLSM are misspecified, which results in poor performance. However, even in this scenario, our proposed method outperforms these models. For $\mathcal M = \{1,2,3\}$, our method combines inter-layer and intra-layer averaging, achieving results similar to $\mathcal M = \{2\}$.

\begin{figure}[H]
	\centering
	\includegraphics[width=6in,height=\textheight]{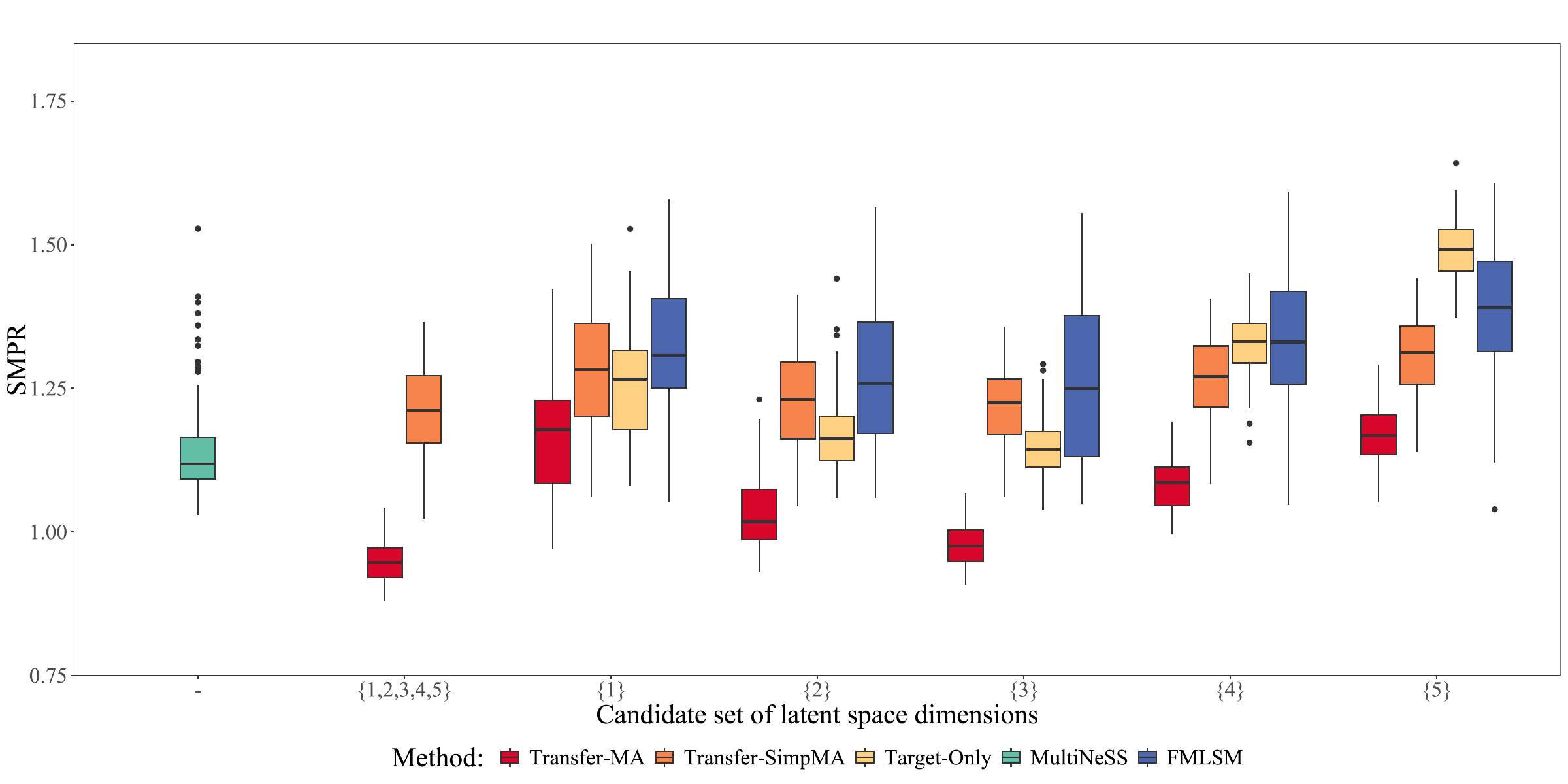}
	\caption{Results of Example 2 (Gaussian Type)} 
	\label{gaussian example2}
\end{figure}

Figure \ref{gaussian example2} displays the results of Example 2. Our method achieves optimal performance when $\mathcal M = \{1,2,3,4,5\}$. For $\mathcal M$ with a single latent vector dimension, Target-Only and FMLSM performance declines as $d_{(1)}$ deviates from the true dimension of the target layer.

\begin{figure}[H]
	\centering
	\includegraphics[width=6in,height=\textheight]{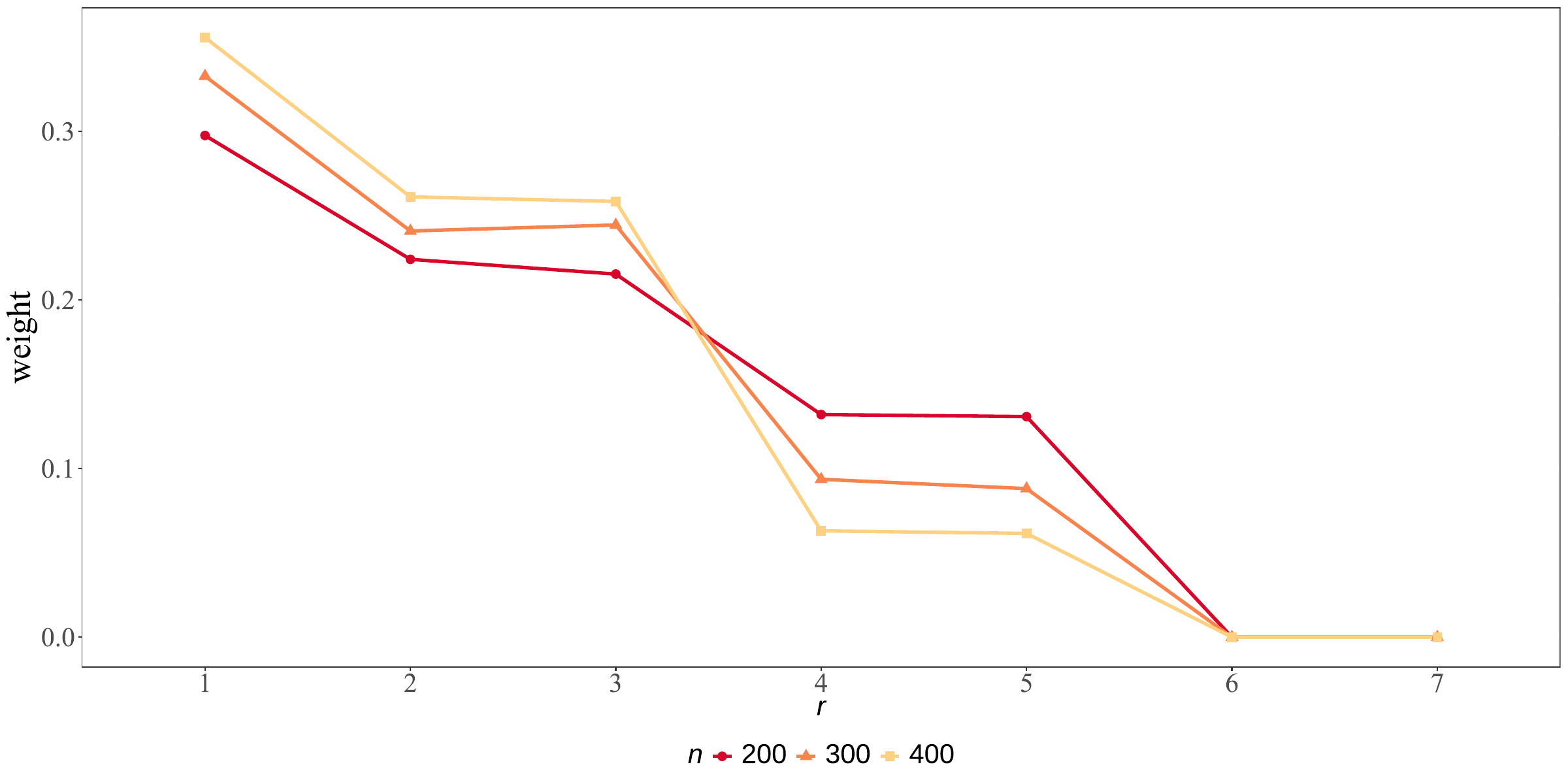}
	\caption{Results of Example 3  } 
	\label{gaussian example3}
\end{figure}

Figure \ref{gaussian example3} presents the results for Example 3. As $n$ grows, the weights assigned to the informative models  ($r=1,2,3$) increase, while the weights of the non-informative models decrease. More specifically, the second and third layers contain some noise, resulting in their weights being lower than that of the target layer. The fourth and fifth layers have less noise compared to the sixth and seventh layers, and thus their weights converge to zero more slowly.

\begin{figure}[H]
	\centering
	\includegraphics[width=6in,height=\textheight]{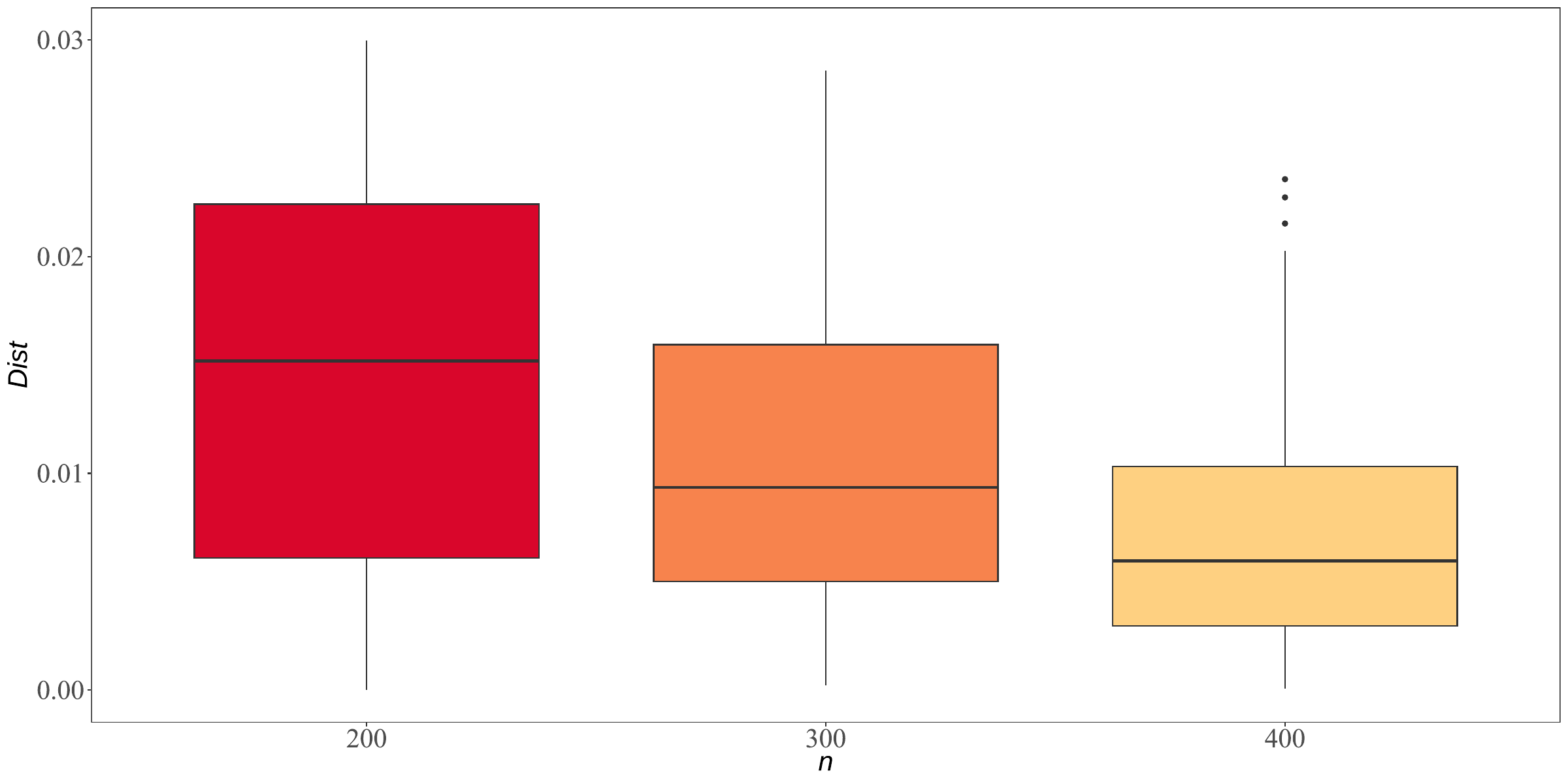}
	\caption{Results of Example 4} 
	\label{gaussian example4}
\end{figure}

Figure \ref{gaussian example4} presents the results of Example 4.   The y-axis represents the distance
$Dist=\inf_{\boldsymbol w\in \underline{\mathcal W}_{\mathcal S^c}}\|\widehat{\boldsymbol w}_{\mathcal S^c}/\|\widehat{\boldsymbol w}_{\mathcal S^c}\|_1-\boldsymbol w_{\mathcal S^c}\|_2=\|\widehat{\boldsymbol w}_{\mathcal S^c}/\|\widehat{\boldsymbol w}_{\mathcal S^c}\|_1-(0.5,0.5)\|_2$, where $\mathcal S^c=\{(2,1),(3,1)\}$. The results show that as  $n$ increases, the normalized weights for the second and third layers converge toward the ideal combination $(0.5,0.5)$, thereby empirically verifying the weight convergence stated in Theorem 2.

\begin{figure}[H]
	\centering
	\includegraphics[width=6in,height=\textheight]{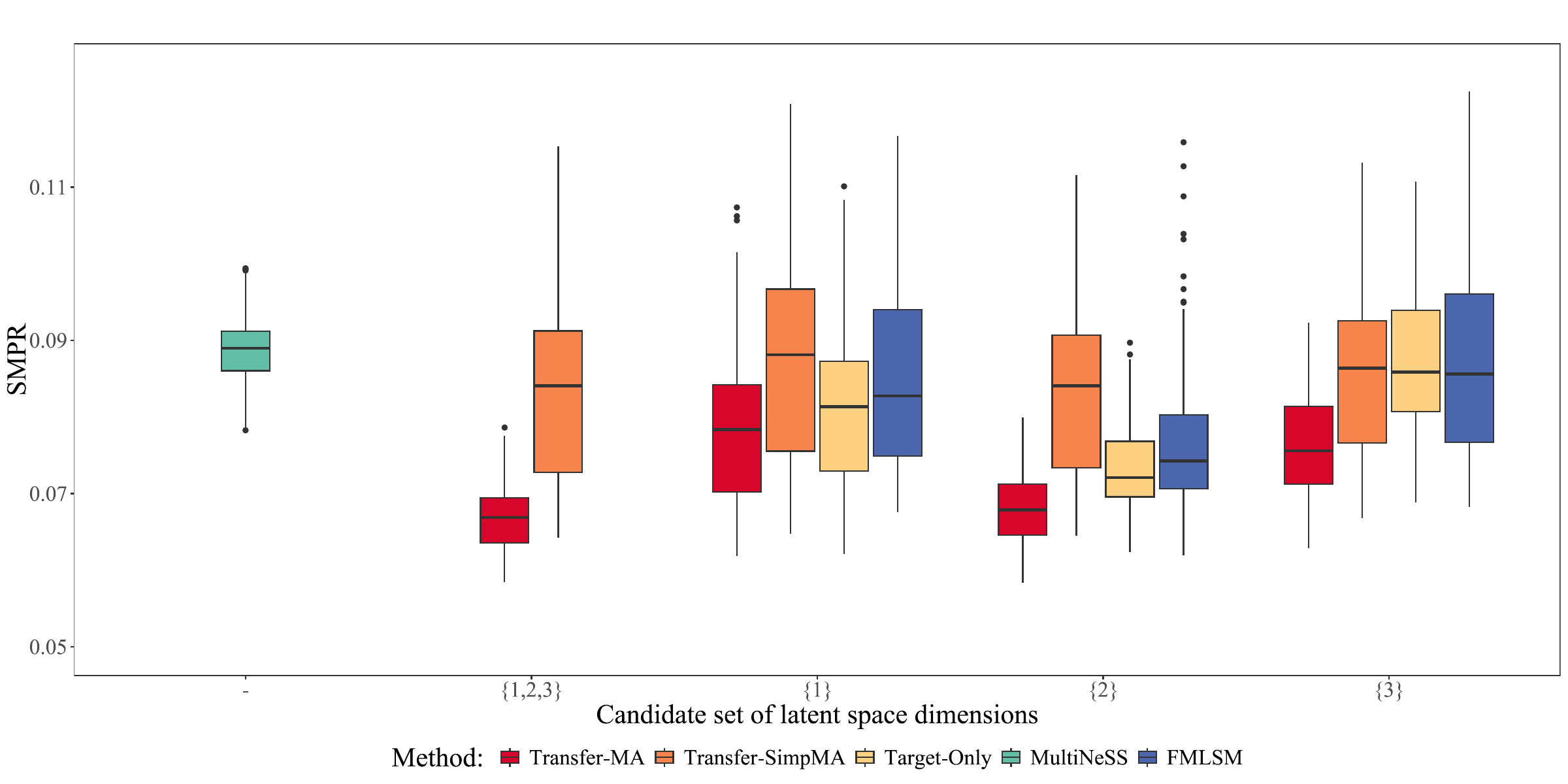}
	\caption{Results of Example 5} 
	\label{example5}
\end{figure}

Figure \ref{example5} presents the results of Example 5. Compared to Figure \ref{gaussian example base1}, all models show reduced performance due to the working model being misspecified relative to the true data-generating process. We observe that $d_{(m)} = 2$ appears to be a ``more correct'' dimension, as Transfer-SimpMA, Target-Only, and FMLSM perform best with $d_{(m)} = 2$ compared to other dimensions. However, for $\mathcal M = \{1,2,3\}$, our method achieves the best performance among all models, suggesting that a single model cannot capture all available information and that model averaging within the layer is necessary.
\begin{figure}[H]
	\centering
	\includegraphics[width=6in,height=\textheight]{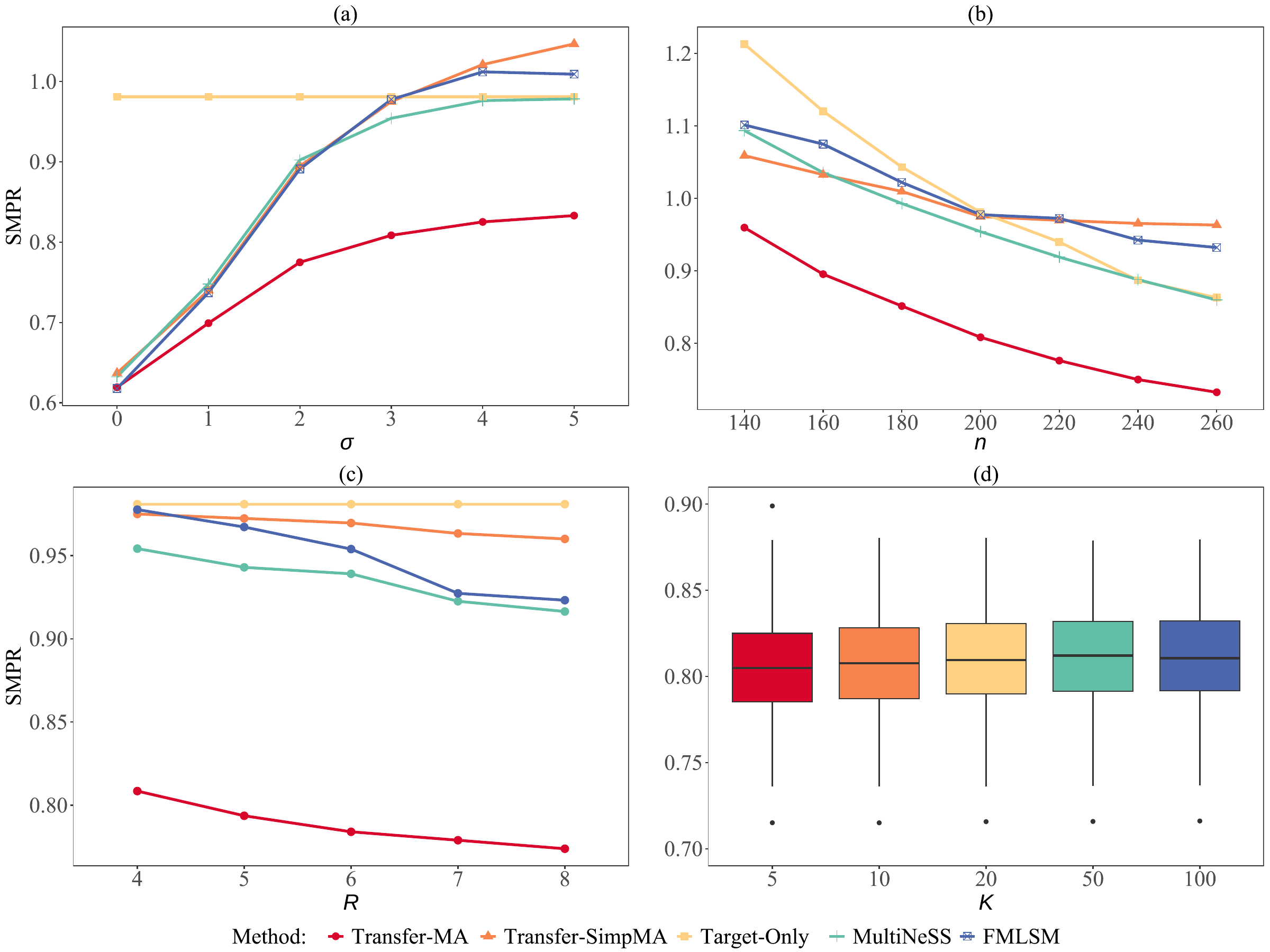}
	\caption{Result of Example 1: Case (a)-(d) (Gaussian Type)} 
	\label{gaussian example1}
\end{figure}

\section{Real Data}\label{real data}
In this section, we apply our proposed method to two real-world social network datasets.
\subsection{Predicting Social Relationships Among Aarhus University Employees}
The first dataset is a multiplex social network collected from the Department of Computer Science at Aarhus University, Denmark \citep{magnani2013combinatorial}. It includes 61 employees and captures their relationships across five different online and offline interactions: Facebook connections, leisure activities, work collaborations, co-authorships, and lunch meetings. We represent each employee as a node and construct a five-layer multilayer network, where  $A_{r,ij}\in \{0,1\}$ for $r=1,2,3,4,5$ and $1\le i<j\le 61.$

In each layer, we randomly select 25\% of the edges as missing. We treat each layer as the target layer in turn and evaluate the predictive performance of different methods on the missing edges. Since the true values ${\mathbb E}(A_{r,ij})$ are unknown, we use the square of mean prediction error (SMPE) as the evaluation metric:  $\text{SMPE}=\sqrt{\sum_{(i,j)\in \mathcal G/\mathcal G_r}\left(\widehat{A}_{r,ij}(\boldsymbol w)-A_{r,ij}\right)^2}$ as the evaluation metric.

\begin{figure}
	\centering
	\includegraphics[width=6in,height=\textheight]{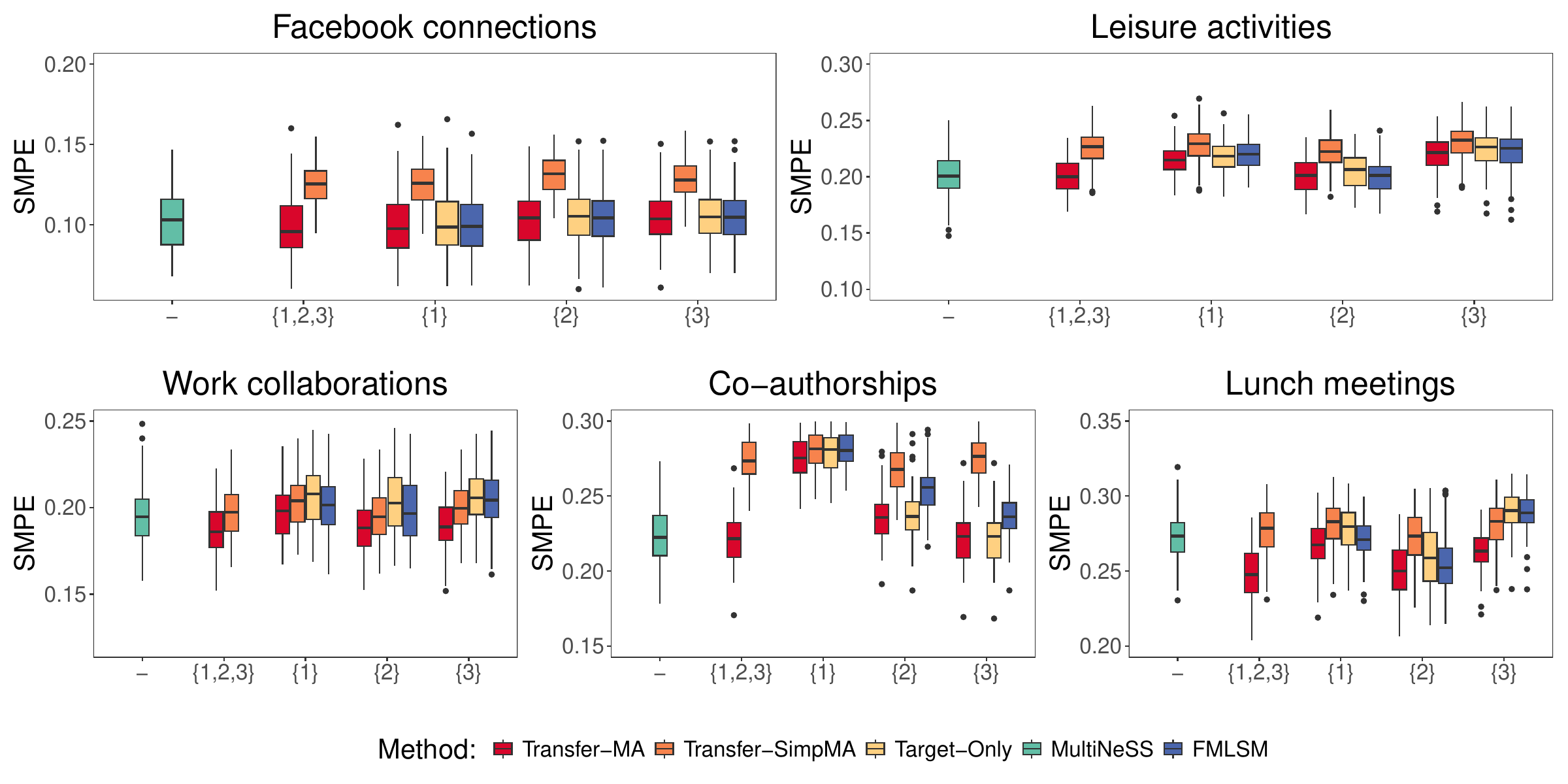}
	\caption{Boxplots of SMPE for Five Layers in the Aarhus Dataset } 
	\label{cs}
\end{figure}

In this case, we set $\mathcal M\in\{\{1\},\{2\},\{3\},\{1,2,3\}\}$ and $K=10$. Figure \ref{cs} shows boxplots depicting the results for each layer. The results reveal significant differences in latent vectors across the five-layer network, as FMLSM exhibits negative performance gains compared to Target-Only. MultiNeSS performs well on the second and third layers but poorly on the other layers. In contrast, our proposed method consistently outperforms FMLSM, Target-Only, and Transfer-SimpMA when $\mathcal M$ contains only one candidate dimension and achieves the best performance when $\mathcal M=\{1,2,3\}$.

\subsection{Predicting User Friendship Relationships Across Two Social Platforms}
This dataset comprises two network layers collected from the social networking platforms Facebook and Twitter \citep{du2019joint}. Each node represents a user with accounts on both platforms, and edges indicate friendships between users. We randomly select $n \in \{100, 300, 500\}$ users as nodes and set 25\% of the edges as missing. Similar to the first dataset, we treat each layer as the target layer in turn and evaluate the predictive performance of different methods on the missing edges.

\begin{figure}
	\centering
	\includegraphics[width=6in,height=\textheight]{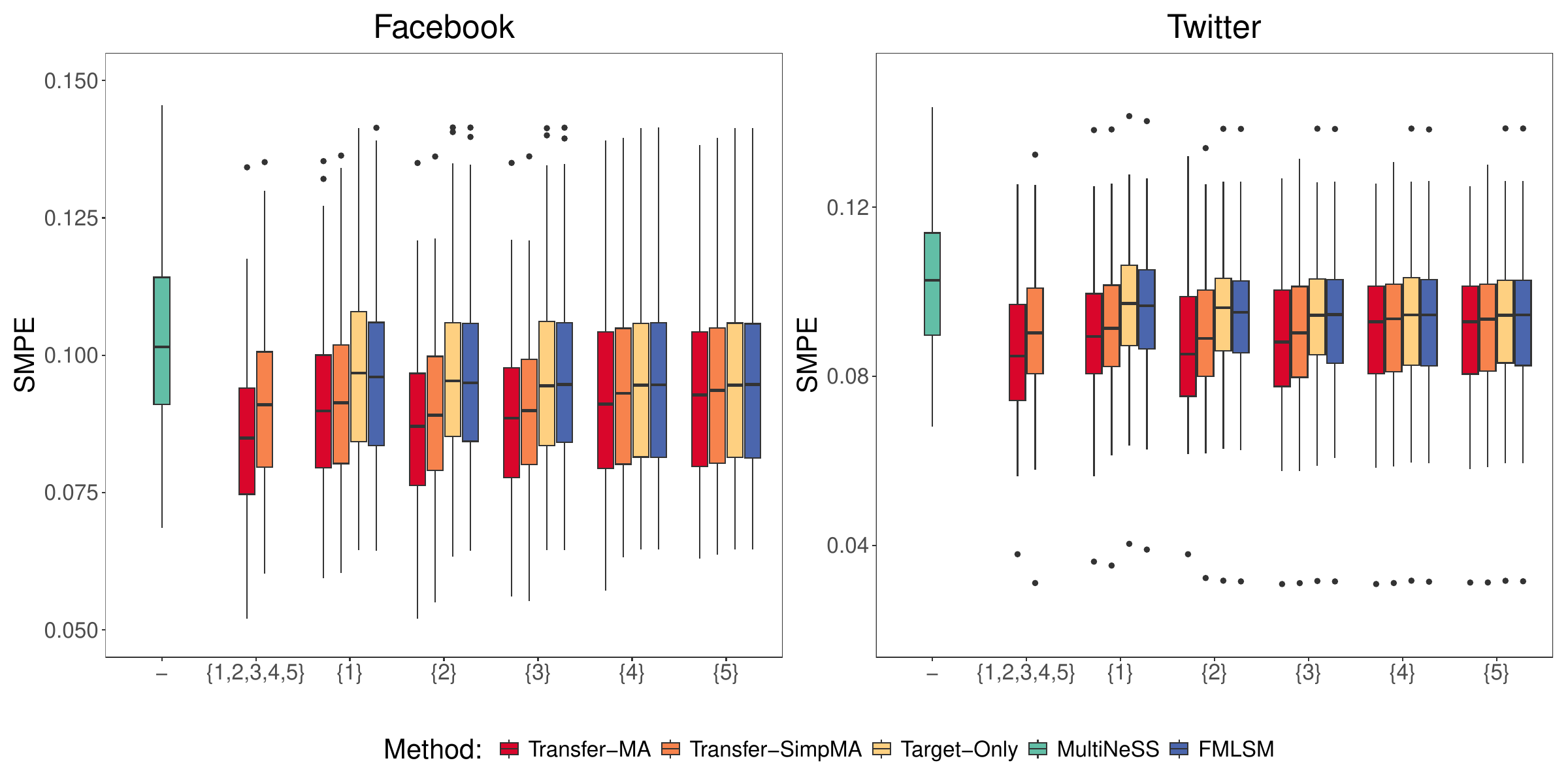}
	\caption{Boxplots of SMPE for Two Layers in the Facebook-Twitter Dataset ($n=100$)} 
	\label{yt100}
\end{figure}

\begin{figure}
	\centering
	\includegraphics[width=6in,height=\textheight]{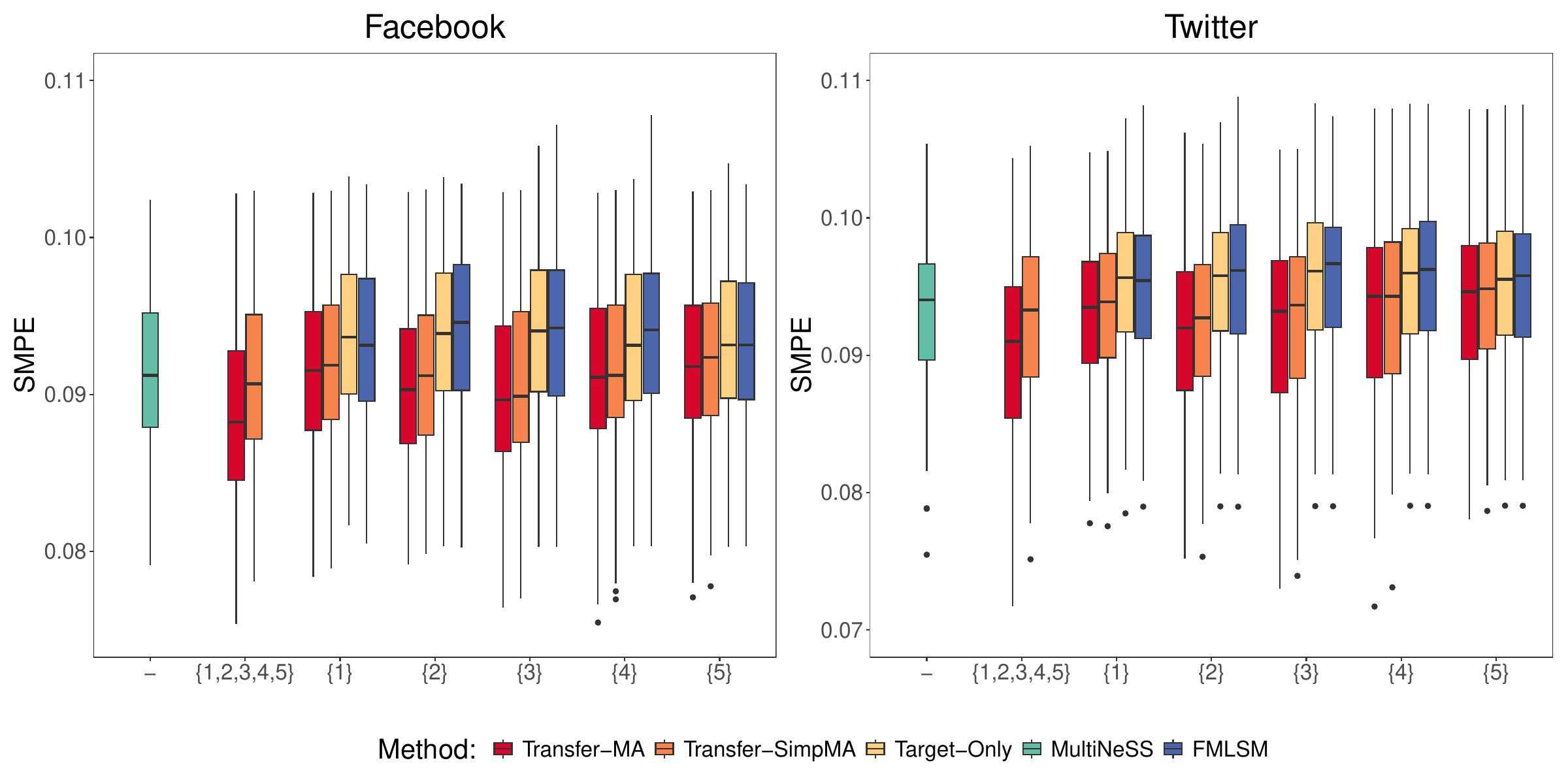}
	\caption{Boxplots of SMPE for Two Layers in the Facebook-Twitter Dataset ($n=300$)} 
	\label{yt300}
\end{figure}

\begin{figure}
	\centering
	\includegraphics[width=6in,height=\textheight]{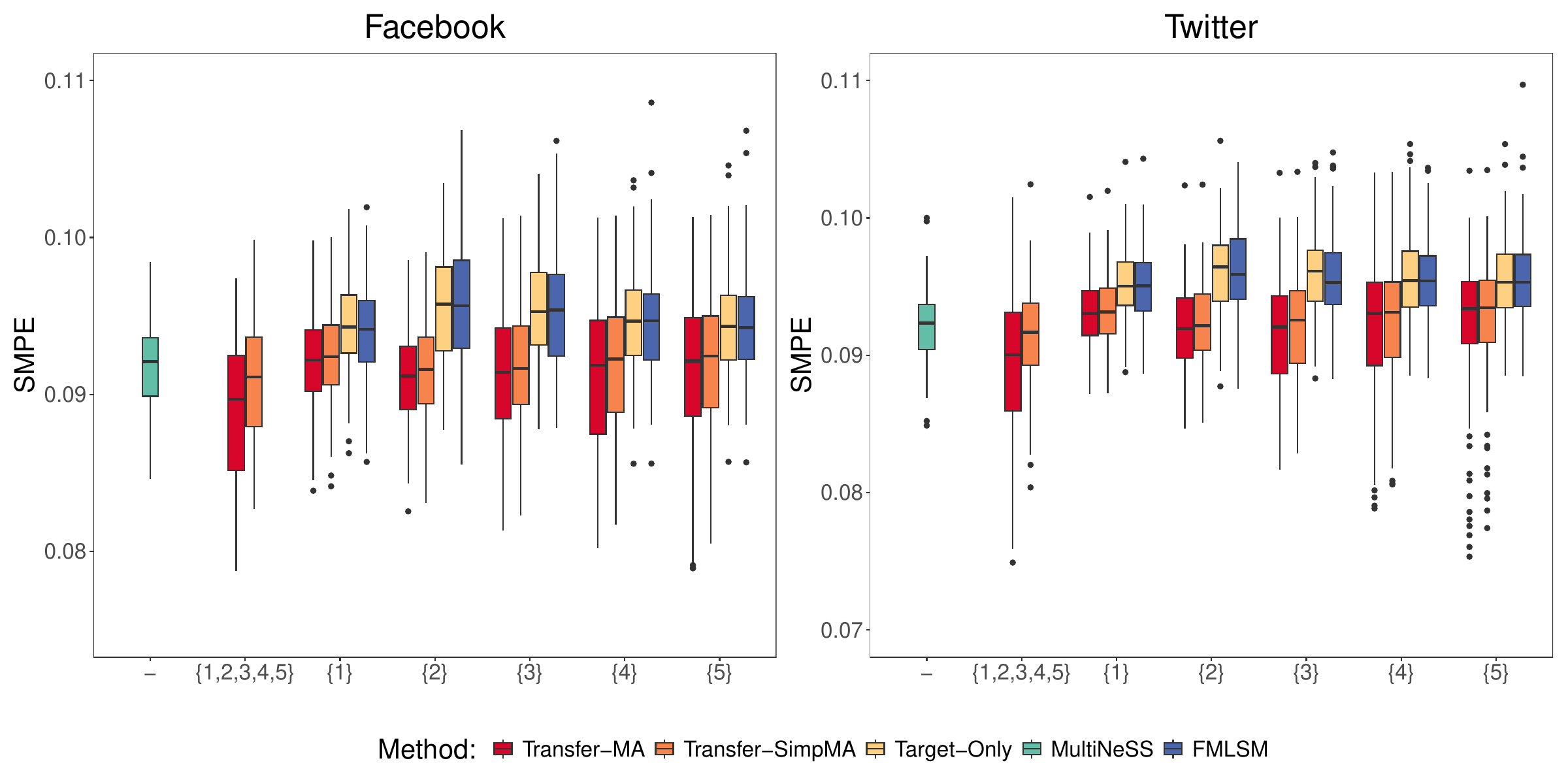}
	\caption{Boxplots of SMPE for Two Layers in the Facebook-Twitter Dataset ($n=500$)} 
	\label{yt500}
\end{figure}

Here, we set $\mathcal M \in \{\{1\},\{2\},\{3\},\{4\},\{5\},\{1, 2, 3, 4, 5\}\}$ and $K=10$. Figures \ref{yt100}–\ref{yt500} display the results for $n = 100, 300, 500$, respectively. Overall, our method achieves the best performance across all network scales. Specifically, no single dimension is optimal for all layers across network scales, as Target-Only shows similar performance for different values of $d_r$ ($r = 1, 2$). While our method significantly outperforms Target-Only for a single candidate dimension, it achieves the best performance when $\mathcal M = \{1, 2, 3, 4, 5\}$. This highlights the importance of accounting for model uncertainty.

\section{Discussion and Conclusions}\label{conclusion}
Transferring information across multiple networks to enhance link prediction in a target layer is a crucial challenge in real-world applications. In this study, we introduce a transfer learning framework for multilayer networks based on a bi-level model averaging method. Our method eliminates the need for prior knowledge of shared structures across layers, automatically assigning weights to each model in a fully data-driven manner. This approach facilitates information transfer from auxiliary layers while mitigating model  uncertainty, significantly enhancing prediction robustness. When informative models exist among the candidates, our method automatically discards the non-informative ones. If all individual models are non-informative but a weighted combination of them forms an informative model, the estimated weights will converge to such a combination. If no such informative combination exists, it still has the desirable property that the ratio of the prediction risk for all edges to the optimal risk converges to 1.  Furthermore, our method supports distributed estimation for auxiliary layers on separate servers, enabling parallel computing for improved efficiency and eliminating the need for raw data transmission. Through simulation experiments, we evaluated our method's performance across diverse scenarios, including both continuous and binary edges. The results demonstrate that our method effectively transfers information from auxiliary layers and exhibits superior robustness. We further validated our method's effectiveness through applications to two real-world datasets.

Several directions for future research emerge from this work. One promising direction is to explore scenarios where multilayer networks have non-identical nodes across layers. For instance, different social platforms may have partially overlapping user bases, with the user sets not being fully identical. In such cases, information from non-overlapping users could still be valuable for link prediction in the target layer. Developing models to handle such scenarios is an important direction for future research.

  \bibliography{bibliography.bib}

\end{document}